# Graphical Abstract

## A Copula-Guided Temporal Dependency Method for Multitemporal Hyperspectral Images Unmixing

Ruiying Li, Bin Pan, Qiaoying Qu, Xia Xu, Zhenwei Shi

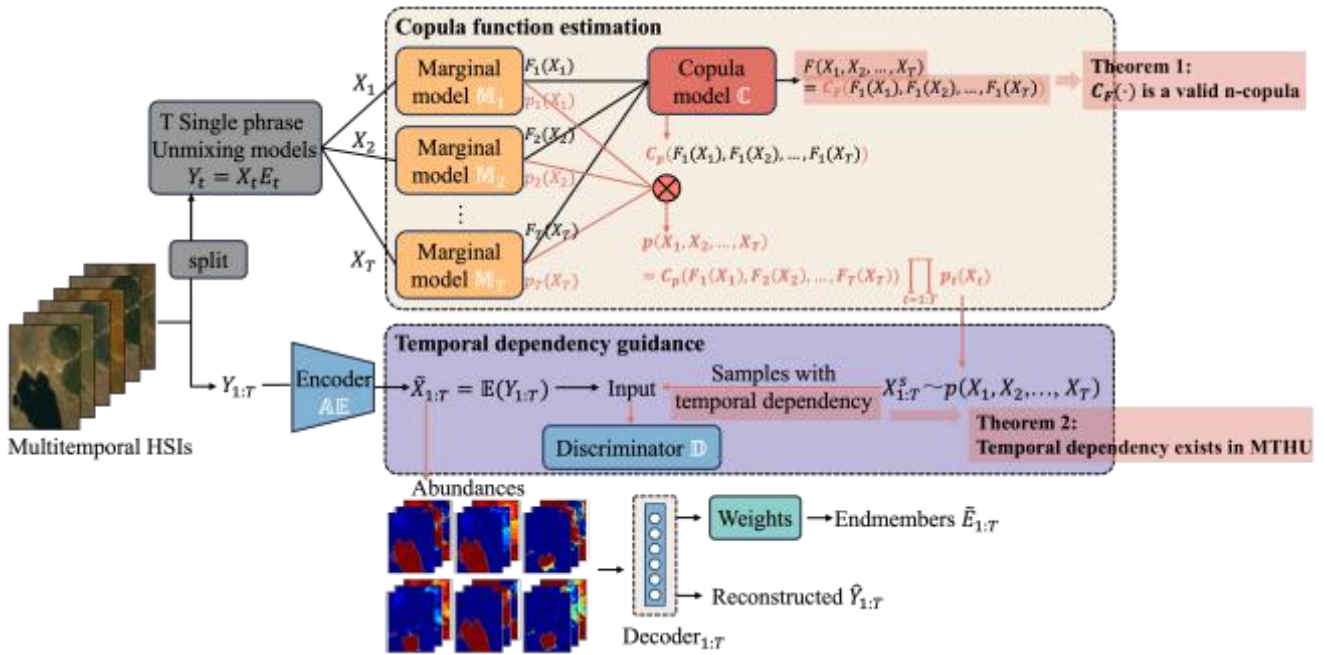

# Highlights

## A Copula-Guided Temporal Dependency Method for Multitemporal Hyperspectral Images Unmixing

Ruiying Li, Bin Pan, Qiaoying Qu, Xia Xu, Zhenwei Shi

- We propose a new problem definition for multitemporal hyperspectral unmixing, which models temporal dependency by copula theory mathematically.

- We solve MTHU problem by developing a copula-guided framework, which estimates dynamical endmembers and abundances with temporal dependency.

- We construct a copula function estimation and a temporal dependency guidance module, which computes and employs temporal dependency, respectively.

- We provide theoretical support, which demonstrates the validity of estimated copula function and the existence of temporal dependency.

# A Copula-Guided Temporal Dependency Method for Multitemporal Hyperspectral Images Unmixing

Ruiying Li[a], Bin Pan[a,*], Qiaoying Qu[a], Xia Xu[b] and Zhenwei Shi[c]

[a] *the School of Statistics and Data Science, KLMDASR, LEBPS, and LPMC, Nankai University, Tianjin, 300071, China*
[b] *the School of Computer Science and Technology, Tiangong University, Tianjin, 300387, China*
[c] *the Image Processing Center, School of Astronautics, and the State Key Laboratory of Virtual Reality Technology and Systems, Beihang University, Beijing, 100191, China*

ABSTRACT

Multitemporal hyperspectral unmixing (MTHU) aims to model variable endmembers and dynamical abundances, which emphasizes the critical temporal information. However, existing methods have limitations in modeling temporal dependency, thus fail to capture the dynamical material evolution. Motivated by the ability of copula theory in modeling dependency structure explicitly, in this paper, we propose a copula-guided temporal dependency method (Cog-TD) for multitemporal hyperspectral unmixing. Cog-TD defines new mathematical model, constructs copula-guided framework and provides two key modules with theoretical support. The mathematical model provides explicit formulations for MTHU problem definition, which describes temporal dependency structure by incorporating copula theory. The copula-guided framework is constructed for utilizing copula function, which estimates dynamical endmembers and abundances with temporal dependency. The key modules consist of copula function estimation and temporal dependency guidance, which computes and employs temporal information to guide unmixing process. Moreover, the theoretical support demonstrates that estimated copula function is valid and the represented temporal dependency exists in hyperspectral images. The major contributions of this paper include redefining MTHU problem with temporal dependency, proposing a copula-guided framework, developing two key modules and providing theoretical support. Our experimental results on both synthetic and real-world datasets demonstrate the utility of the proposed method.

## 1. Introduction

Hyperspectral image (HSI) contains rich spectral information, enabling significant potential in diverse applications, such as agriculture, military and forestryDong, Liu, Du and Zhang(2022b);Xiong, Zhou and Qian(2020);Peng, Xie, Zhao, Wang, Yee and Meng (2020); Li, Shi, Wang, Xi, Li and Gamba (2022). However, accurate materials identification in various fields can be hindered by mixed pixels phenomenon, which arises due to low spatial resolution Dong, Zhou, Wu, Wu, Shi and Li (2021). To enable reliable image interpretation, hyperspectral image unmixing is required, which decomposes mixed pixel by estimating spectral signatures (Endmembers) and their coefficients (Abundances) Borsoi, Imbiriba and Bermudez (2020). The underlying pixel mixing models are typically assumed to be characterized as either linear or nonlinear Bioucas-Dias, Plaza, Dobigeon, Parente, Du, Gader and Chanussot(2012). The linear mixing model (LMM) represents mixed spectrum as a weighted linear combination of endmembers, where the weights correspond to abundances Gao, Han, Hong, Zhang and Chanussot (2022).

LMM-based unmixing models can be categorized into five principal approaches: geometric Nascimento and Dias (2005);Drumetz, Chanussot, Jutten, Ma and Iwasaki(2020a), statistical Zhou and Rodrigues (2024); Zhao, Wang, Chen and Chen (2022), nonnegative matrix factorizationYang, Zhou, Xie, Ding, Yang and Zhang (2011); Dong, Lu, Liu and Yuan (2022a), sparse regressionXu, Pan, Chen, Shi and Li (2021); Ren, Hong, Gao, Sun, Huang and Chanussot (2023) and deep learningTao, Paoletti, Wu, Haut, Ren and Plaza (2024); Hong, Gao, Yao, Yokoya, Chanussot, Heiden and Zhang(2022); Chen, Gamba and Li(2023); Jin, Ma, Fan, Huang, Mei and Ma (2023). These methods are primarily designed for single-phrase unmixing, processing individual image captured at one time frame. However, as sensors can actually acquire images at one fixed location across multiple temporal intervals, individual image can be extended to a sequence of multitemporal images, inherently capturing material dynamics. Naturally, single-phrase unmixing can be expanded to multitemporal hyperspectral unmixing (MTHU), which further investigates the temporal evolution of endmembers and abundancesThouvenin, Dobigeon and Tourneret (2016).

Building upon single-phrase hyperspectral unmixing techniques and multitemporal sequence images, MTHU aims to model the temporal variability of endmembers and the dynamics of abundancesHenrot, Chanussot and Jutten (2016). Recently promising MTHU approaches can be broadly classified into two main categories. The first category methods decouple MTHU as a unmixing process and a temporal dynamics modeling process. Based on explicit expressions, simple dynamics structures are represented through parametric functions or probabilistic frameworks. Such as state-space models, Kalman filters, ordinary differential equations and Bayesian distributions Thouvenin et al.

*Corresponding author.
✉ panbin@nankai.edu.cn ( Bin Pan)
ORCID(s): 0009-0004-3059-9023 ( Bin Pan)

(2016); Henrot et al.(2016); Yokoya, Zhu and Plaza(2017); Sigurdsson, Ulfarsson, Sveinsson and Bioucas-Dias (2017); Borsoi, Imbiriba, Closas, Bermudez and Richard (2022); Drumetz, Mura, Tochon and Fablet (2020b); Li, Pan, Ma, Xu and Shi(2025b); Bhatt, Joshi and Vijayashekhar(2018); Thouvenin, Dobigeon and Tourneret (2018); Liu, Lu, Wu, Du, Chanussot and Wei (2022); Sun and Liu (2023). The second category methods address MTHU problem in data driven approaches. Based on implicit expressions, complex temporal patterns are fitted by deep learning methods, such as transformers and recurrent neural networks Li, Dong, Xie, Xu, Li and Shi (2025a); Borsoi, Imbiriba and Closas (2023). Furthermore, research on multitemporal sequence images, which reveal underlying patterns of material evolution, is critical for monitoring dynamic phenomenaZurita-Milla, Gomez-Chova, Guanter, Clevers and Camps-Valls (2011); Iordache, Tits, Bioucas-Dias, Plaza and Somers (2014); Dutta, Rahman and Kundu (2015); Gudex-Cross, Pontius and Adams (2017); Bullock, Woodcock and Olofsson (2020); Laamrani, Joosse, McNairn, Berg, Hagerman, Powell and Berry (2020); Wang, Ding, Tong and Atkinson (2021); Kathirvelu, Yesudhas and Ramanathan (2023); Wang, Ding, Tong and Atkinson (2022). The dynamical patterns can be explicitly learned by modeling the joint distribution and dependency structure of image data.

While the previous categories show promising results, current dynamical functions and neural networks fail to capture the temporal dependency structure of material evolution jointly. To address these limitations, MTHU methods should explore the explicit expression for complex temporal dependency in hyperspectral image sequences.

Copula theory is a promising approach to model temporal dependency structure, which offers a flexible and interpretable framework. Since copula functions enable the derivation of dependency structures directly from marginal distributionsNelsen(2006); Joe(2014), copula may provide a mathematical framework to simultaneously represent dynamical systems and capture temporal dependency. Initially proposed in statistics, copula theory links marginal distributions in order to simulate joint distribution of random variablesDurante and Sempi (2016); Schweizer and Wolff (1981); Sklar (1959). Gradually, copula theory has been extended to various fieldsXu and Cao (2024); Ozdemir, Allen, Choi, Wimalajeewa and Varshney (2018), including hyperspectral image processing for tasks such as image classification, change detection and image fusionLi, Li, Wang and Varshney (2023); Tamborrino and Mazzia (2023);Li, Wang, Li, Geng and Varshney(2025c). Given the high dimensional of hyperspectral images, recent advances demonstrate that powerful neural networks offer promising solutions for estimating copula functions Tagasovska, Ackerer and Vatter (2019); Janke, Ghanmi and Steinke (2021); Drouin, Marcotte and Chapados (2022); Zeng and Wang(2022); Ashok, Marcotte, Zantedeschi, Chapados and Drouin (2024). Inspired by previous studies, this research aims at exploring a mathematical framework to model temporal dependency in MTHU, guided by the principles of copula theory.

However, directly integrating copula into MTHU presents several challenges. First, existing copula-based methods model dependency without accounting for spectral characteristics of multitemporal hyperspectral images. Second, guiding unmixing process through temporal dependency lacks a unified mathematical model and framework. Additionally, the validity of estimated temporal copula function is not verified.

In order to address above issues, in this paper, we propose a **Co**pula **g**uided **T**emporal **D**ependency multitemporal unmixing (Cog-TD) method, which describes new MTHU mathematical model, constructs a copula-guided framework, and presents two key modules supported by theoretical demonstration. To explicitly define MTHU problem, we propose a mathematical model, which models temporal dependency structure by utilizing copula density function. To guide MTHU process by temporal dependency, we develop a copula-guided framework, which estimates dynamical evolution of endmembers and abundances. To compute and employ temporal dependency, we design a copula function estimation module and a temporal dependency guidance module. To support the proposed method theoretically, we demonstrate validity and existence theorem, which describes that the estimated copula function is valid and temporal dependency exists in MTHU. The main contributions of proposed method can be summarized as follows:

1. We propose a new problem definition for multitemporal hyperspectral unmixing, which models temporal dependency by copula theory mathematically.
2. We solve MTHU problem by developing a copula-guided framework, which estimates dynamical endmembers and abundances with temporal dependency.
3. We construct a copula function estimation and a temporal dependency guidance module, which computes and employs temporal dependency, respectively.
4. We provide theoretical support, which demonstrates the validity of estimated copula function and the existence of temporal dependency.

The remainder of this paper is structured as follows. Section II describes the related work. Section III introduces the proposed algorithm and theoretical properties. Section V describes the experimental results on both synthetic datasets and real datasets.

## 2. Background
### 2.1. Multitemporal Hyperspectral Unmixing

Considering that sequences of hyperspectral images $Y_{1:T}$ are acquired at fixed location across $T$ time frame. At single time $t \in 1 : T$, linear mixing model is adopted generally, which represents mixed pixel in $Y_t$ as a linear combination of endmembers, that is

$$Y_t = X_t E_t + N_t \tag{1}$$

where $Y_t \in \mathbb{R}^{H \times W \times C}$ is hyperspectral image acquired at time $t$. Assume that the number of endmembers is $P$, then $X_t \in \mathbb{R}^{H \times W \times P}$ is abundance and $E_t \in \mathbb{R}^{P \times C}$ is endmember. Building linear mixing model and incorporat- ing temporal information, endmembers and abundances in MTHU are estimated temporally different. The dynamics of materials can be extracted by temporal functions Thouvenin et al. (2016); Henrot et al. (2016); Yokoya et al. (2017); Sigurdsson et al. (2017); Borsoi et al. (2022); Drumetz et al. (2020b); Li et al. (2025b); Bhatt et al. (2018); Thouvenin et al. (2018); Liu et al. (2022); Sun and Liu (2023) or deep learning methods Li et al. (2025a); Borsoi et al. (2023). The temporal functions are expressed as:

$$\begin{cases} X_t = f_X(X_{t-1}) \\ E_t = f_E(E_{t-1}) \end{cases} \quad (2)$$

where function $f_X$ represents dynamics of abundance and $f_E$ denotes variability of endmembers. Both $f_X$ and $f_E$ are modified according to the specific construction.

For example, existing methods consider $f_X$ as linear transformation Thouvenin et al. (2016), statistical model Borsoi et al. (2022); Bhatt et al. (2018); Thouvenin et al. (2018); Liu et al. (2022); Sun and Liu (2023) and ordinary differential equation Henrot et al. (2016); Li et al. (2025b):

$$X_t = X_{t-1} + \epsilon_t \quad (3)$$
$$X_t \sim P(X_t | X_{t-1}) \quad (4)$$
$$\frac{dX_t}{dt} = G(X_t; y_t; t) \quad (5)$$

For $f_E$, following perturbed linear mixing model, some methods model variability of endmember as an additive perturbation Thouvenin et al. (2016, 2018); Liu et al. (2022); Borsoi et al. (2023). While statistical model Bhatt et al. (2018); Sun and Liu (2023) and ordinary differential equation Drumetz et al. (2020b) are also considered, thus $f_E$ has the following expressions:

$$E_t = E + dE_t \quad (6)$$
$$E_t \sim P(E_t | E_{t-1}) \quad (7)$$
$$\frac{dE_t}{E_t} = G(E_t; t) \quad (8)$$

Previous MTHU methods unmix pixels based on Eq.(1), and model temporal dynamics of materials by functions $f_X$ and $f_E$. These methods consider unmixing process and dynamics modeling as actually two divided structures, which are slightly linked. Another approach estimate end- members and abundances by deep learning methods, such as transformer and recurrent neural network Li et al. (2025a); Borsoi et al. (2023). However, dynamical functions and deep learning methods are both insufficient for extracting temporal information.

In order to further extract temporal information in images sequences and explore multitemporal unmixing process, this paper aims to model temporal dependency by copula theory.

## 2.2. Copula Theory

From one point a view, copulas are functions that join or "couple" multivariate distribution functions to their one-dimensional marginal distribution functions Joe (2014). Alternatively, copulas are multivariate distribution functions whose one-dimensional margins are uniform on the interval (0,1).

The word copula means "a link", which is originally from Latin noun Nelsen (2006). Copula is the function that links multivariate joint distribution function to their marginal distribution functions. Then, copula is related to the study of dependence among random variables in statistics Schweizer and Wolff (1981). For constructing multivariate joint distribution function, a $T-$dimensional copula function $C_F : [0, 1]^T \rightarrow [0, 1]$ is employed, which enhances to construct both dependency structure and joint distribution from estimated marginal distribution functions.

Formally, according to Sklar's theorem Sklar(1959), the joint cumulative distribution function (CDF) of any random vector $[X_1, X_2, ..., X_T]$ can be expressed as a combination of a copula $C_F$ and the marginal CDF $F(X_i) = P(X_i \leq x_i)$. Sklar's theorem can be demonstrated by equation as:

$$F(X_1, X_2, ..., X_T) = P(X_1 \leq x_1, X_2 \leq x_2, ..., X_T \leq x_T)$$
$$= C_F(F(x_1), F(x_2), ..., F(x_T)) \quad (9)$$

where $F(X_1, X_2, ..., X_T)$ is joint cumulative distribution function, $F(X_i) = P(X_i \leq x_i)$ is marginal cumulative distribution function and $C_F(.)$ is copula function.

Every CDF corresponds to a probability density function (PDF). For example, marginal $F(X_i)$ corresponds to $p(X_i) = P(X_i = x_i)$, copula function $C_F(.)$ corresponds to $C_p : [0, 1]^T \rightarrow [0, 1]$, and joint $F(X_1, X_2, ..., X_T)$ corresponds to $p(X_1, X_2, ..., X_T)$ that can be also described by $p(X_i)$ and $C_p(.)$.

Given the high dimension of hyperspectral images, estimating copula function $C_F(.)$ and $C_p(.)$ by deep learning methods is promising. In which Neural Copula Zeng and Wang (2022) estimates the marginal distributions and copula function by solving differential equations with complex constraints. Specifically, Neural Copula Zeng and Wang (2022) considers the boundary of copula. We denote marginal CDF as $F(X_i) \doteq u_i$. Let $u_j^0$ express that $u_i = 0$ while other $u_j$ is normal. Let $u_j^1$ express that $u_i = u_i$ while other $u_j = 1$. For example, $u_1^0 = [0, u_2, ..., u_d]$, $u_d^1 = [1, 1, ..., 1, u_d]$. Then boundary conditions can be demonstrated as:

$$\begin{cases} C_F(u_1^0) = 0; C_F(u_2^0) = 0; ...; C_F(u_{d-1}^0) = 0 \\ C_F(u_1^1) = u_1; C_F(u_2^1) = u_2; ...; C_F(u_d^1) = u_d \end{cases} \quad (10)$$

Boundary conditions are also satisfied in our method since we estimate copula based on Eq.(10). However, incorporating copula into MTHU faces several difficulties: copula-based methods lack spectral information, guidance framework is not constructed, and validity of copula is not guaranteed.

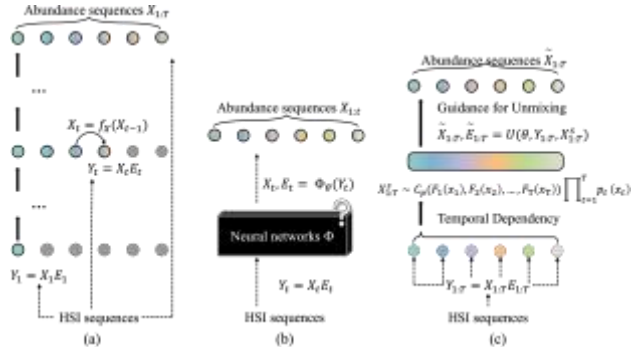

**Figure 1:** The sketch of different demonstrations for temporal dynamics in MTHU, which is modeled by (a) dynamical function $f_X$ ; (b) neural networks; (c) temporal dependency (our method). We explore the explicit expression by temporal dependency to consider the joint distribution across different phrases.

## 3. Methodology

In this section, the proposed method is described in detail. In Section 3.1, we propose new copula-based problem definition for MTHU, which constructs mathematical temporal dependency model by developing copula theory. In Section 3.2, we demonstrate the proposed overall framework, which includes copula function estimation module (Section 3.3) and temporal dependency guidance module (Section 3.4). In Section 3.5, comprehensive analysis about theoretical support is presented.

### 3.1. Copula–based Problem Definition for MTHU

In this subsection, in order to model dynamical structure by explict expressions, we define temporal dependency based on copula theory and formulate MTHU as new mathematical model, that is Eq.(15).

Previous MTHU methods model dynamical structure by temporal functions $f_X$ or deep learning approaches, which are described in Fig.1 (a) and Fig.1 (b), respectively. Both dynamical functions and deep learning methods are insufficient for extracting temporal information. Different from previous researches, this paper aims to demonstrate dynamical structure as temporal dependency by copula theory explicitly, which is described in Fig.1 (c).

We consider hyperspectral sequence images $Y_{\{1:T\}}$ and single image $Y_t \in R^{H \times W \times C}$. According to Eq.(1), every $Y_t$ corresponds to a single-phrase abundance $X_t \in R^{H \times W \times P}$, which can be obtained from normal unmixing method:

$$\text{for } t \text{ in } [1:T]; X_t = s_t(Y_t) \quad (11)$$

where $s_t$ is single-phrase model. $X_t$ can be flatten as a vector $X_t \in R^{HWC \times 1}$. Assume that $X_t$ is a sample belonging to a probability density function (PDF) $p_t(X_t)$, that it $X_t \sim p_t(X_t)$. Then $X_{\{1:T\}}$ belongs to a joint distribution $p(X_1, X_2, ..., X_T)$.

In the field of statistics, variables are independent when equation $p(X_1, X_2, ..., X_T) = \prod_{t=1}^{T} p_t(x_t)$ holds. However,

as $X_{\{1:T\}}$ is dependent with each other temporally, there exists unknown variables dependency. The relationship between joint distribution and dependent marginals is described as:

$$p(X_1, X_2, ..., X_T) \neq \prod_{t=1}^{T} p_t(x_t) \qquad (12)$$

Existing in multiple temporal frames, the unknown variables dependency describes correlation between material evolution, which is demonstrated as temporal dependency.
In order to derive the unknown temporal dependency, we develop copula theory, which expresses joint distribution as copula function and marginal distributions. According to Sklar's theorem (Eq.(9)), joint CDF $F(X_1; X_2; ...; X_T)$ can be expressed as a combination of copula function $C_F(.)$ and marginal CDF $F_t(X_t)$. As copula PDF $C_p(.)$ can be obtained from the derivation of $C_F(.)$, the joint PDF can be estimated as:

$$p(X_1; X_2; ...; X_T) = P(X_1 = x_1; X_2 = x_2; ...; X_T = x_T)$$
$$= C_p(F_1(x_1), F_2(x_2), ..., F_T(x_T)) \times \prod_{i=1}^{T} p_i(x_i)$$
$$(13)$$

where $C_p : [0, 1]^T \to [0, 1]$ is the density function of copula function $C_F$ in Eq.(9). Compared Eq.(13) with Eq.(12), the reason why equation holds is the value of $C_p$. Actually, $C_p$ can describe the dependency between marginals and joint distributions. The independence of random vector $\{X_1; X_2; ...; X_T\}$ equals to $C_p = 1$. On the contrary, when $C_p(.) \neq 1$, temporal abundance $X_{\{1:T\}}$ is dependent with each other. In this case, temporal dependency is regarded as existing in multitemporal hyperspectral images. Thus $C_p$ can capture complicated correlations between variables.

As joint distribution $p(X_1; X_2; ...; X_T)$ can be computed by Eq.(13), drawn sample from $p(X_1; X_2; ...; X_T)$ can obtain samples $X^s_{1:T}$ with temporal dependency. It should be noted that $X^s_{1:T}$ is not abundance, it is merely feature map carrying temporal dependency. In order to guide MTHU by temporal dependency, we impose the distribution of sample $X^s_{1:T}$ on the distribution of abundances $X_{1:T}$, that is:

$$p(X_{1:T}) = \int p(X_{1:T}|X^s_{1:T})p(X^s_{1:T})dX^s_{1:T} \qquad (14)$$

where $p(X^s_{1:T})$ is sample distribution and $p(X_{1:T})$ is the unknown abundance distribution. $p(X_{1:T}|X^s_{1:T})$ represents the encoder in an AE framework and the generative model in an adversarial framework. Under the guidance of $X^s_{1:T}$, MTHU can capture the dynamics of multitemporal abundances and endmembers. MTHU model is then formulated as:

$$\begin{cases} X^s_{1:T} \sim C_p(F_1(x_1), F_2(x_2), ..., F_T(x_T)) \prod_{t=1}^{T} p_t(X_t) \\ X_{1:T} \sim \int p(X_{1:T}|X^s_{1:T})p(X^s_{1:T})dX^s_{1:T} \\ Y_{1:T} = X_{1:T} E_{1:T} \end{cases}$$

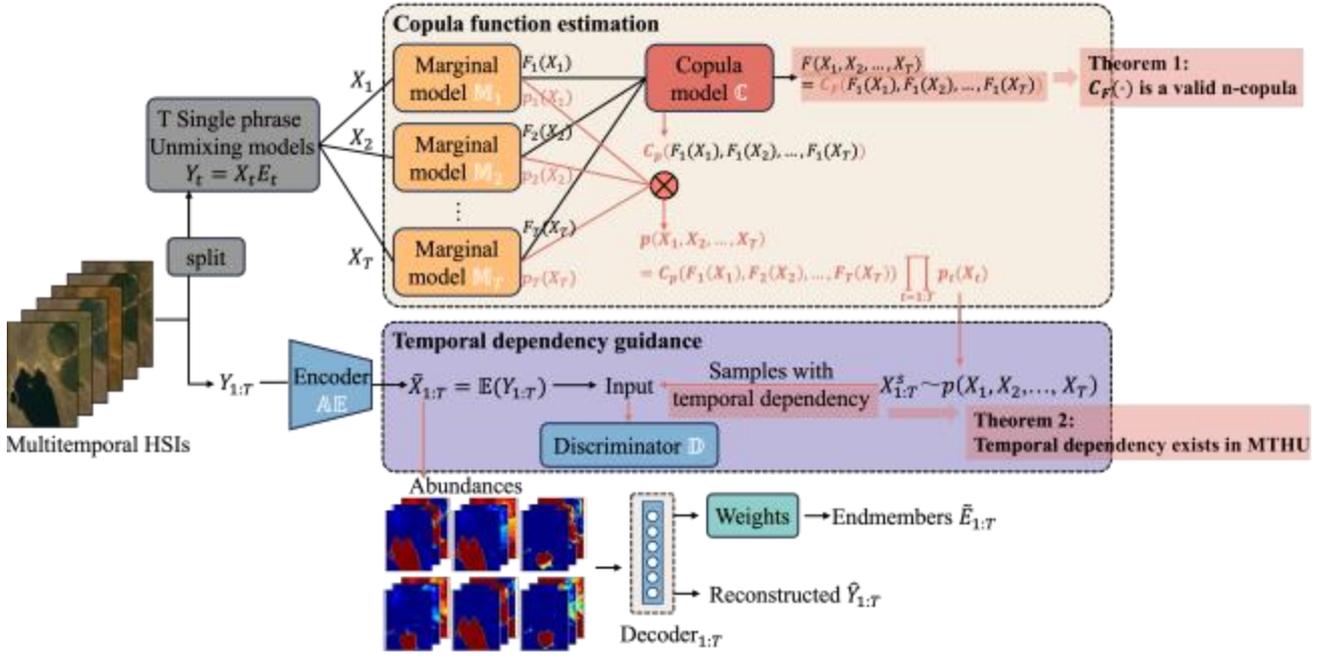

**Figure 2:** The overall framework of proposed algorithm. We take multitemporal images $Y_{1:T}$ as input, which are processed by both copula function estimation module and temporal dependency guidance module. Multitemporal abundance $X_{1:T}$ and endmembers $E_{1:T}$ then can be obtained as output.

(15)

where $Y_{1:T}$ is hyperspectral sequence, $X_{1:T}$ and $E_{1:T}$ are estimated abundances and endmembers with temporal dependency. $X^s_{1:T}$ is sampled vectors with temporal dependency. These variables can be computed by overall multitemporal unmixing framework, which is introduced in following Section 3.2.

### 3.2. Overall Multitemporal Unmixing Framework

In this subsection, we propose copula guided temporal dependency multitemporal unmixing framework to estimate and utilize temporal dependency. This framework including copula function estimation module and temporal dependency guidance module, which are introduced in Section 3.3 and Section 3.4 respectively.

As described in Fig.2, the proposed framework $U(\ )$ takes multitemporal images sequence $Y_{1:T}$ as input. We first estimates the temporal dependency structure using the copula function estimation module. The estimated temporal dependency is then utilized by the temporal dependency guidance module to guide unmixing process. The outputs of overall framework are the multitemporal endmembers $E_{1:T}$ and abundances $A_{1:T}$, which incorporate the temporal dependency captured by the copula model. The whole framework is shown as:

$$X_{1:T}; E_{1:T} = U(\ ; Y_{1:T})$$
$$\text{where } U(\ ) = M_t(\ _M) \cup C(\ _C) \cup AE(\ _{AE}) \cup D(\ _D)$$ (16)

where $M_t(\ _M) \cup C(\ _C)$ represent copula estimation module and $AE(\ _{AE}) \cup D(\ _D)$ represent temporal dependency guidance module.

In order to estimate marginal distributions and compute copula function, copula estimation module constructs marginal model and copula model, which is expressed as:

$$p(X^s_{1:T}) = C(\ _C; M_t(\ _M; Y_{1:T}))$$ (17)

where $p(X^s_{1:T})$ is predicted joint distribution, sequence $Y_{1:T}$ is input of copula estimation module, $M_t(\ _M)$ is marginal model and $C(\ _C)$ is copula model. We first split sequence $Y_{1:T}$ as individual image $Y_t$ to obtain single-phrase abundance $X_t$. Then single-phrase abundances $[X_1; X_2; ....; X_T]$ are viewed as inputs of marginal models, which output estimated marginal distribution $p_t(X_t)$ with CDF $F_t(X_t)$. Finally, input $F_t(X_t)$ to copula model for the estimation of joint distribution $p(X_1; X_2; ....; X_T)$.

In order to guide the prediction process of temporal abundances and endmebers, temporal dependency guidance module develops autoencoder network and discriminator network, which is expressed as:

$$X_{1:T}; E_{1:T} = D(\ _D; AE(\ _{AE}; Y_{1:T}; X^s_{1:T}))$$ (18)

where $X_{1:T}; E_{1:T}$ are endmembers and abundances with temporal dependency. Sample $X^s_{1:T}$ is drawn from $p(X^s_{1:T})$ in Eq.(17), sequence $Y_{1:T}$ is input of autoencoder network. $AE(\ _{AE})$ is autoencoder network and $D(\ _D)$ is discriminator network. We extract latent vectors from encoder of $AE(\ _{AE})$ and drawn samples with temporal dependency from joint distribution. Then latent vectors and samples are taken as inputs of discriminator, this operation can impose temporal guidance on vectors in latent space. Finally, latent vectors with temporal dependency are abundances and the weights of decoders are endmembers.

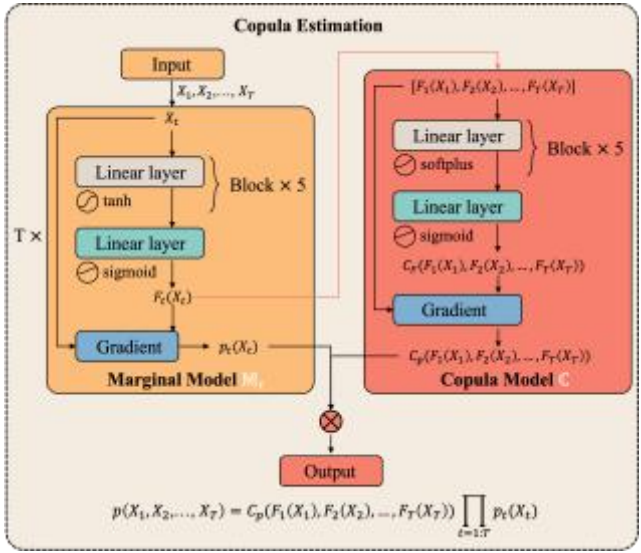

**Figure 3:** The framework of copula estimation module. We take single-phrase abundances as inputs. After Marginal Model processing, marginal distributions can be obtained, which will be delivered to Copula Model. After computing, copula function representing temporal dependency is acquired. Then, both marginal distributions and copula are multiplied to get joint distribution.

The copula estimation module calculates copula function and joint distribution by neural networks simultaneously. Copula function represents temporal dependency, and samples with temporal dependency can be sampled from joint distribution. The temporal dependency guidance module takes common autoencoder, but imposes temporal guidance on vectors in latent space, which further improve the multitemporal unmixing process. More detailed introduction about copula estimation part and temporal dependency guidance part is show in Section 3.3 and Section 3.4 respectively.

### 3.3. Copula Function Estimation Module

As temporal dependency is associated with function $C_p(.)$, in this section, we present the estimation method for copula functions $C_f(.)$ and $C_p(.)$.

This copula function estimation module includes two kinds of networks: $T$ marginal models $\mathbf{M}_t(\theta_\mathbf{M})$ and one copula model $\mathbf{C}(\theta_\mathbf{C})$. $\mathbf{M}_t(\theta_\mathbf{M})$ is developed to estimate marginal distributions, which is the foundation of $\mathbf{C}(\theta_\mathbf{C})$, model for computing copula function. Finally, joint distribution can be derived according to Eq.(13). The whole estimation process can be described as follows:

$$\text{for every } t \in [1:T] \; F_t(X_t); p_t(X_t) = \mathbf{M}_t(\theta_\mathbf{M}; X_t) \quad (19)$$

$$C_f(F(x_1); F(x_2); ...; F(x_T)); C_p(F(x_1); F(x_2); ...; F(x_T)) = \mathbf{C}(\theta_\mathbf{C}; [F_1(X_1); F_2(X_2); ...; F_T(X_T)]) \quad (20)$$

$$p(X^s_{1:T}) = C_p(F(x_1), F(x_2), ..., F(x_T)) \times \prod_{i=1}^{T} p_i(x_i) \quad (21)$$

The framework of copula estimation part is shown in Fig.3. It consists of $T$ marginal models $\mathbf{M}_t$ with parameter $\theta_\mathbf{M}$ and a copula model $\mathbf{C}$ with parameter $\theta_\mathbf{C}$. Abundance $X_t$ of each phrase is input of every corresponding marginal model. In which linear layer and tanh function construct a block, and five blocks form hidden layer. For output layer, features are passed through linear layer and sig- moid function to estimate the marginal CDF value $F_t(X_t)$. Moreover, marginal PDF value $p_t(p_t)$ can be computed at the same time, which is achieved by gradient layer, with automatic derivation mechanism. Then all CDF values $[F_1(X_1); F_2(X_2); ...; F_T(X_T)]$ are fed into the copula model to obtain copula function value $C_f(.)$. Different from Neural Copula model, we improve hidden layers in copula model by developing linear layer and softplus function as one block, and five blocks construct hidden layer. Then the output $h_{i+1}$ of $i+1$th layer can be expressed as:

$$h_{i+1} = \ln(1 + e^{W_i h_i + b_i}); i \in [0; ...; 4] \quad (22)$$

For output layer, features are passed through linear layer and sigmoid function to estimate the copula CDF value $F_t(X_t)$. After gradient layer, copula PDF can be obtained, which decides the existence of temporal dependency. Thus copula CDF $C_f(F(x_1); F(x_2); ...; F(x_T))$ and copula PDF $C_p(F(x_1); F(x_2); ...; F(x_T))$ are obtained as Eq.(20).

According to Eq.(19), Eq.(20) and Eq.(21), the joint distribution $p(X_1; X_2; ...; X_d)$ can be computed. Drawn samples from it can obtain $X^s_{1:T}$ which reflects temporal dependency.

The loss functions of marginal model are composed of four parts: fitness $L^\mathbf{M}_1(\theta^\mathbf{M}_1)$, non-negative $L^\mathbf{M}_2(\theta^\mathbf{M}_2)$, integral-to-one $L^\mathbf{M}_3(\theta^\mathbf{M}_3)$ and boundary conditions $L^\mathbf{M}_4(\theta^\mathbf{M}_4)$, which can be formulated as:

$$L_\mathbf{M}(\theta_\mathbf{M}) = \sum_{i=1}^{4} \lambda_i L^\mathbf{M}_i(\theta^\mathbf{M}_i) \quad (23)$$

where

$$\begin{cases} L_1^M(M_1) = -\frac{1}{T}\sum_{i=1}^{T}\left[\log P_t(X_t)\right] \\ L_2^M(M_2) = \frac{1}{T^2}\sum_{i=1}^{T^2} relu(-P_t(X_t)) \\ L_3^M(M_3) = \frac{1}{T}\sum_{i=1}^{T}\left|1 - sum(F_t(X_t))\right| \\ L_4^M(M_4) = \sum_{i=1}^{T}(F_t(0) + |1 - F_t(1)|) \end{cases} \quad (24)$$

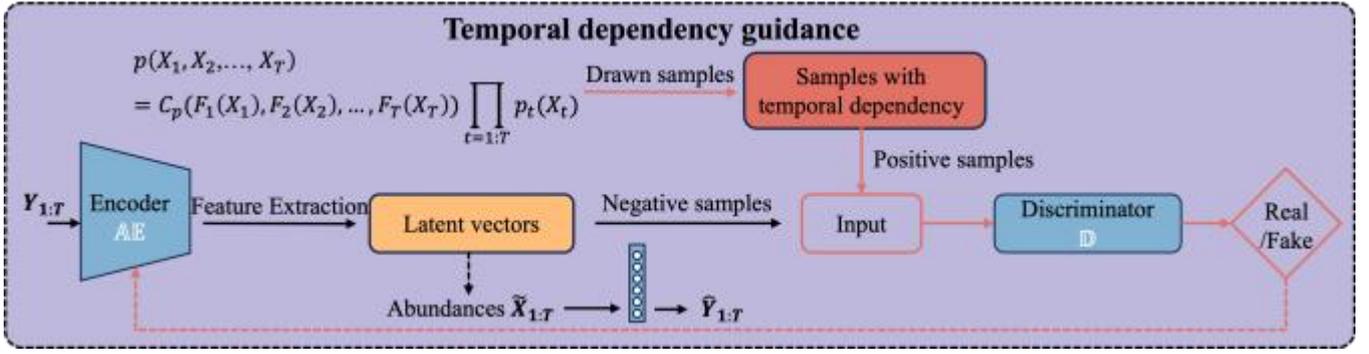

**Figure 4:** The framework of temporal dependency guidance. Samples with temporal dependency will be drawn from joint distribution. Latent vectors are extracted from Encoder. Discriminator take samples as positive samples and take latent vectors as negative samples, which aims at imposing temporal dependency on abundances.

The loss functions of copula model are composed of five parts: fitness $L_1^C(\theta_1^C)$, non-negative $L_2^C(\theta_2^C)$, integral-to-one $L_3^C(\theta_3^C)$, boundary conditions $L_4^C(\theta_4^C)$ and observation constraint $L_5^C(\theta_5^C)$. Following Eq.(10), we set $F_i^0$ express that $F_i = 0$ while other $F_j$ is normal. Let $F_i^1$ express that $F_i = F_i$ while other $F_j = 1$ which can be formulated as:

$$L_C(\theta_C) = \sum_{i=1}^{5} \lambda_i L_i^C(\theta_i^C) \qquad (25)$$

where

$$\begin{cases} L_1^C(\theta_1^C) = -sum\left[\log C_p(F_1, F_2, ..., F_T)\right] \\ L_2^C(\theta_2^C) = \frac{1}{T} sum(-relu(C_p(F_1, F_2, ..., F_T))) \\ L_3^C(\theta_3^C) = \left|1 - sum(C_F(F_1, F_2, ..., F_T))\right| \\ L_4^C(\theta_4^C) = \sum_{i=1}^{T}\left(C_p(F_i^0) + \left|F_i - C_p(F_i^1)\right|\right) \\ L_5^C(\theta_5^C) = \left|C_F(F_1, F_2, ..., F_T) - F_{observed}(X_1, X_2, ..., X_T)\right| \end{cases} \qquad (26)$$

### 3.4. Temporal Dependency Guidance Module

In this section, we express the proposed guidance strategy for unmixing by temporal dependency, which is described in Fig.4.

The main purpose of temporal dependency guidance part is imposing temporal prior on the latent vectors, which is extracted from encoder. In section 3.3, temporal dependency is estimated in the form of distribution $C_p(.)$, which is regarded as temporal prior in this section. We draw samples from $C_p(.)$ to acquire $X_{1:T}^s$ which is shown as features with temporal dependency. The temporal dependency guidance part includes two kinds of networks: autoencoder $AE(\theta_{AE})$ and discriminator $D(\theta_D)$. $AE(\theta_{AE})$ is developed to extract latent vectors from hyperspectral sequence and reconstruct images. $D(\theta_D)$ is designed to transfer temporal information into latent vectors by incorporating adversarial training. The whole guidance process can be described as follows:

$$\text{sample } X_{1:T}^s \text{ from } C_p(F(x_1); F(x_2); ...; F(x_T)) \times \prod_{i=1}^{T} p_i(x_i) \qquad (27)$$

$$X_{1:T}; E_{1:T} = AE(\theta_{AE}; Y_{1:T}; D(\theta_D; X_{1:T}^s)) \qquad (28)$$

The framework of guidance part is shown in Fig.4. Autoencoder $AE(\theta_{AE})$ is developed to predict abundances and endmembers. The encoder is used for extracting abundances, and the linear decoder is used to reconstruct the images. Decoder will be initialized by VCA method, which is also adopted as reference endmember constraints.

The discriminator $D(\theta_D)$ is trained to distinguish the sample with temporal dependency (positive samples) that we are trying to model or latent vectors from encoders (negative samples). As a result, networks update parameters in iterations to learn abundances with temporal dependency.

According to Eq.(27) and Eq.(28), the dynamical abundances $X_{1:T}$ and endmembers $E_{1:T}$ are estimated. According to them can reconstruct the original hyperspectral image sequence. The loss functions of guidance part are composed of three parts: adversarial $L_G(\theta_G)$, $L_{AE}(\theta_{AE})$ and endmember $L_E$, which can be formulated as:

$$L_{guidance} = L_G(\theta_G) + L_{AE}(\theta_{AE}) + L_E \qquad (29)$$

The solutions to the min–max adversarial autoencoder can be expressed as:

$$L_G(\theta_G) = \min_{E} \max_{D} E\left[\log \mathbb{D}(X_{1:T}^s)\right] + E\left[\log(1 - X_{1:T})\right] \qquad (30)$$

The reconstruction loss is:

$$L_{AE}(\theta_{AE}) = \|Y_{1:T} - \hat{Y}_{1:T}\|_2 \\
= \|Y_{1:T} - U(\theta; X_{1:T}; E_{1:T}; X_{1:T}^s)\|_2 \qquad (31)$$

The endmember loss is:

$$L_E = \| E_{1:T} - E_{reference} \|_2 \tag{32}$$

## 3.5. Theoretical Analysis

In this section, we provide validity theorem and existence theorem. Validity theorem shows that solving Eq.(15) by proposed algorithm can lead to a valid copula. Existence theorem indicates that the value of $C_p(.)$ represents temporal dependency and temporal dependency exists in MTHU.

### 3.5.1. Validity of Copula

In this subsection, we ensure that the estimation function $C_F(.)$ satisfies the mathematical definition of a copula function.

**Definition 1 (Copula Function):** An n-copula is a function from $I^n$ to $I$ with the following properties:

1. For every $u \in I^n$, $C(u) = 0$ is at least one coordinate of $u$ is 0. And if all coordinates of $u$ are 1 except $u_k$, then $C_u = u_k$.

2. For every $a = [a_1; a_2; ...; a_n] \in I^n$, $b = [b_1; b_2; ...; b_n] \in I^n$ such that $a \leq b$, $V_C([a;b]) \geq 0$, where $V_C([a;b])$ are C-volume of $[a;b]$. Equivalently, the C-volume of an n-box $[a;b]$ is the nth order difference of C on B $[a;b]$. Thus $V_C([a;b])$ is given by:

$$V_C([a;b]) = \Delta_a^b C(x) = \Delta_{a_n}^{b_n} \Delta_{a_{n-1}}^{b_{n-1}} ... \Delta_{a_1}^{b_1} C(x) \tag{33}$$

where

$$\Delta_{a_k}^{b_k} C(x) = C([a_1; a_2; ...; b_k; ...; a_T]) - C([a_1; a_2; ...; a_k; ...; a_T]) \tag{34}$$

**Theorem 1 (Copula Validity):** The function $C_F(.)$ estimated in unmixing process is a valid n-copula.

*Proof:* We consider the definition and properties of n-copula. The first property, that is Eq.(10) actually, is satisfied by constraint $L_4^C(\underset{4}{C})$ in Eq.(26).

For the second property, we set an n-dim vector $x = [x_1; x_2; ...; x_T] \in I^T$ and $x_i \in [a_i; b_i]$. The $C_F$-volume of $[a;b]$ is the nth order difference on $x$, $V_{C_F}([a;b])$ can be expanded as:

$$V_{C_F}([a;b]) =$$
$$\Delta_{a_T}^{b_T} \Delta_{a_{T-1}}^{b_{T-1}} ... \Delta_{T_2}^{T_2} [C_F([b_1; a_2; ...; a_T]) - C_F([a_1; a_2; ...; a_T])]$$
$$= ...$$
$$= \frac{\partial^d C_F(X)}{\partial x_1 \partial x_2 ... \partial x_d} = \frac{\partial^d C_c}{\partial F_1 \partial F_2 ... \partial F_d} \tag{35}$$

then look back on the process in copula model $\mathbb{C}$.

*Input layer:* $F_1; F_2; ...; F_T \doteq h_0 \tag{36}$

*Hidden layers:* $h_{i+1} = \ln(1 + e^{W_i h_i + b_i}) \tag{37}$

*Output layer:* $C_F(X) = \frac{1}{1 + e^{-(W_L h_{L-1} + b_L)}} \tag{38}$

where the derivation of function $\ln(1+e^x)$ and the derivation of function $\frac{1}{1+e^{-(x)}}$ are greater than or equal to 0. As $W_j \geq 0$, $Wh_j + b_j$ is also non-decreasing. Thus the forward process is n-increasing. According to the chain rule, we can obtain that

$$\frac{\partial^d \mathbb{C}_c}{\partial F_1 \partial F_2 \ldots \partial F_d} = V_{C_F}([a,b]) \geq 0 \qquad (39)$$

Thus two properties of copula function are satisfied. Then function $C_F$ is a T-copula.

### 3.5.2. Existence of Temporal Dependency

In this section, we show that the value of $C_p(.)$ reflects temporal dependency and temporal dependency exists in MTHU.

**Theorem 2 (Temporal Dependency Existence):** The temporal dependency can be represented by $C_p(.)$ and temporal dependency exists in MTHU.

*Proof:* As mentioned in Section 3.1 that $C_p(.) \neq 1$ holds when temporal abundance $X_{\{1:T\}}$ is dependent with each other. In this case, temporal dependency exists in MTHU. Thus, the value of $C_p(.)$ can represent whether temporal dependency exists.

Based on previous analysis, we conclude that:

Let abundance $X_t \sim p_t(X_t)$, with its CDF $F_t(X_t)$ as marginal distribution. Then copula PDF $C_p(.)$ can be computed from marginals. The value of $C_p(.)$ indicates the existence of temporal dependency, that is:

$$\begin{cases} \text{Temporal Dependency is not exist, when } C_p(.) = 1 \\ \text{Temporal Dependency exists, when } C_p(.) \neq 1 \end{cases} \qquad (40)$$

Thus, the value of function $C_p(.)$ can represent whether temporal dependency exists. After experiment, we computed that $C_p(.) \neq 1$. Thus temporal dependency exists in MTHU.

## 4. Experiment

In this section, we describe the experimental setup used to evaluate the proposed multitemporal hyperspectral unmixing method. The experiments are designed to test the performance using both synthetic and real datasets. This section includes the settings, dataset descriptions, comparison methods, and evaluation metrics, followed by quantitative and qualitative results to understand the capability completely.

### 4.1. Settings and Datasets

This subsection outlines the experimental settings used for the unmixing tasks. Additionally, we provide a description of the datasets employed, both synthetic and real, to assess the effectiveness of proposed multitemporal hyperspectral unmixing model.

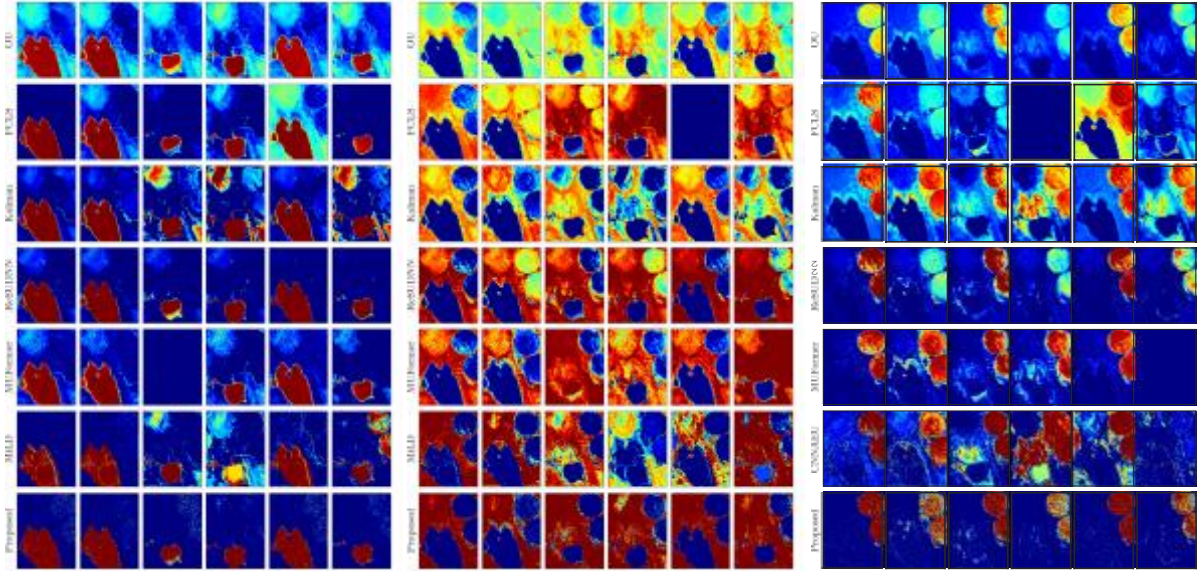

**Figure 5:** Comparison of abundance maps on Lake Tahoe dataset. (Top to bottom) Abundance maps obtained by groundtruth, FCLS, OU, Kalman, ReSUDNN, Muformer, MiLD and proposed method. The top row represents the true abundances of six phrases. (Left to right) Multitemporal abundance maps with water, soil, and vegetation endmembers. Abundances change from phrase 1 to 6.

### 4.1.1. Comparision Methods

To demonstrate the effectiveness of our approach, we compare it with several state-of-the-art MTHU methods. These methods include traditional techniques as well as recent deep learning-based approaches, including those utilizing recurrent neural networks and transformers. The comparison aims to highlight the advantages of our method in handling multitemporal hyperspectral data.

There are six different methods adopted as comparison methods. Fully constrained least squares (FCLS), online unmixing (OU) formulate the problem as a two-stage stochastic programThouvenin et al. (2016), kalman represent the multitemporal mixing process using a state-space formulationBorsoi et al. (2022), recurrent hyperspectral unmixing with neural networks (ReSUDNN) formulate MTHU as a Bayesian inference problemLiu et al. (2022), multitemporal hyperspectral image unmixing transformer (MUFormer) is an end-to-end unsupervised deep learning modelLi et al. (2025b) and multitemporal latent dynamical (MiLD) unmixing frameworkBorsoi et al. (2023). The initialization endmembers of the above methods are obtained by VCA algorithm, which is also adopted to obtain reference endmembers $E_{reference}$.

### 4.1.2. Datasets Description

This subsection provides detailed information on the datasets used for evaluating the proposed multitemporal hyperspectral unmixing approach. The datasets include synthetic data, designed to simulate controlled environments, and real-world hyperspectral images, which reflect the complexities of actual applications. The characteristics of these

**Table 1**

Qualitative Results of FCLS, OU, Kalman, ReSUDNN, MU-Former, MiLD and proposed on Synthetic dataset 1.

| method | $NRMSE_A$ | $NRMSE_E$ | $SAM_E$ | $NRMSE_Y$ |
|---|---|---|---|---|
| FCLS | 0.537 | - | - | 0.086 |
| OU | 0.434 | 0.342 | 0.260 | 0.059 |
| kalman | 0.356 | 0.124 | 0.076 | 0.061 |
| ReSUDNN | 0.318 | 0.117 | 0.075 | 0.088 |
| MUFormer | 0.255 | 0.110 | 0.059 | 0.079 |
| Mild | 0.234 | 0.113 | 0.122 | **0.057** |
| Proposed | **0.138** | **0.109** | **0.055** | 0.064 |

datasets are outlined to give a comprehensive understanding of their suitability for the task.

Lake Tahoe is adopted as real dataset, which is acquired between 2014 and 2015. It consists of $T = 6$ hyperspectral images representing different time phases. Each image in the sequence has dimensions of $N = 150 \times 110$ pixels and $L = 173$ bands remain after preprocessing. There are $P = 3$ predominant endmember classes: Water, Soil, and Vegetation. Notable changes are observed over the lake surface and within agricultural (crop circle) areas across the six acquisition times. Each endmember class displays distinct temporal change patterns.

Synthetic dataset 1 consists of $T = 6$ temporal hyperspectral images. Each image has $N = 50 \times 50$ pixels and $L = 224$ spectral bands. This dataset contains $P = 3$ distinct endmembers, the reference spectra were selected from the USGS library. For the initial time instant (t=1), spatial endmember variability was introduced. The sequence of abundance maps features localized abrupt changes occurring soil, road). To achieve realistic spectral variability, the endmembers for each pixel and each time instant was randomly selected from a library of manually extracted pure pixels representing the four materials, sourced from the Jasper Ridge hyperspectral image. The sequence of abundance maps was generated randomly based on a Gaussian random field and incorporates small, spatially compact abrupt changes throughout the temporal sequence. Hyperspectral images were generated using the linear mixing model and white Gaussian noise was added to achieve a signal-to-noise ratio (SNR) of 30 dB.

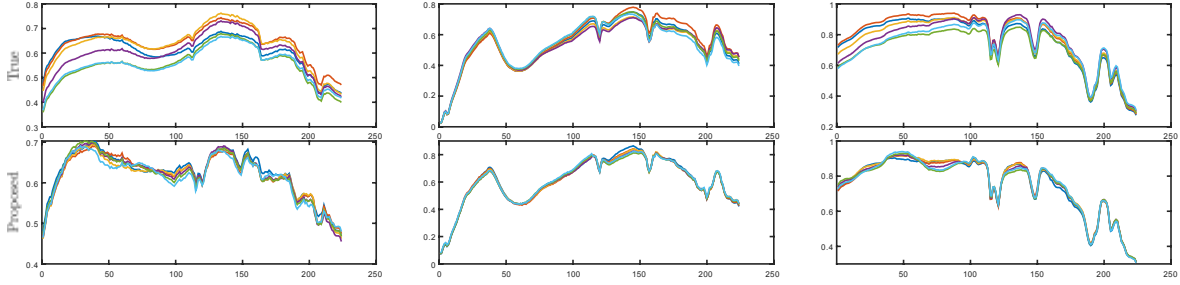

**Figure 6:** The endmember estimation results on Synthetic data 1. The top row represents the true endmember results and the bottom row is the result of the proposed method. Different color lines describe multiple phrases. (Left to right) Endmember 1, 2 and 3.

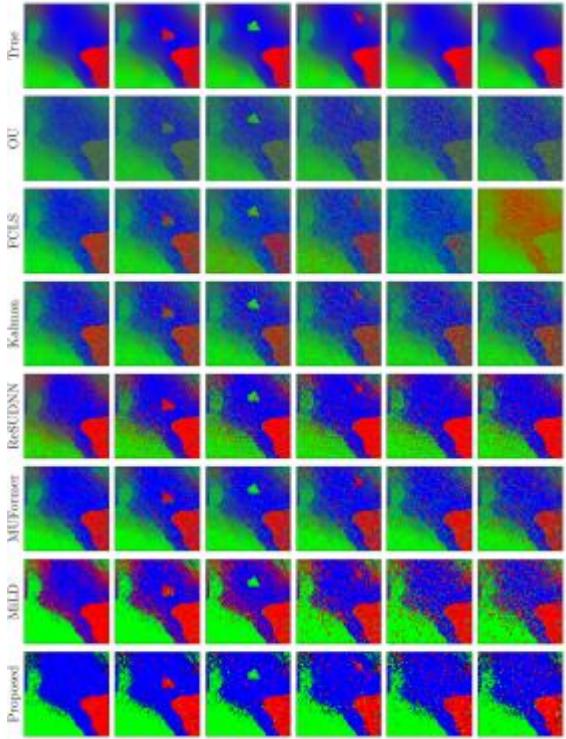

**Figure 7:** Comparison of abundance maps on Synthetic data 1. (Top to bottom) Abundance maps obtained by groundtruth, FCLS, OU, Kalman, ReSUDNN, Muformer, MiLD and proposed method. The top row represents the true abundances of six phrases. (Left to right) Abundance maps obtained from moment 1 to 6. Green, blue, and red represent the three endmembers, respectively.

#### 4.1.3. Evaluation Metrics

To evaluate the performance of the unmixing methods, we employ a variety of quantitative metrics. These include the normalized root mean square error (NRMSE) for abundanaces $NRMSE_A$, endmembers $NRMSE_E$ and reconstructed images $NRMSE_Y$. And the spectral angle mapper (SAM) for endmembers $SAM_E$. These metrics help assess the accuracy of the estimated abundances and the spectral signatures of the endmembers in the context of multitemporal hyperspectral unmixing. They are computed as:

$$NRMSE_A = (\frac{1}{T}\sum_{t=1}^{T}\sum_{n=1}^{N}\frac{\|a_{n,t}-\hat{a}_{n,t}\|^2}{\|a_t\|^2})^{\frac{1}{2}} \quad (41)$$

$$NRMSE_E = (\frac{1}{NT}\sum_{t=1}^{T}\sum_{n=1}^{N}\frac{\|e_{n,t}-\hat{e}_{n,t}\|^2}{\|e_{n,t}\|^2})^{\frac{1}{2}} \quad (42)$$

$$NRMSE_Y = (\frac{1}{T}\sum_{t=1}^{T}\sum_{n=1}^{N}\frac{\|y_{n,t}-\hat{a}_t\hat{E}_{n,t}\|^2}{\|y_{n,t}\|^2})^{\frac{1}{2}} \quad (43)$$

$$SAM_E = \frac{1}{TNP}\sum_{t=1}^{T}\sum_{n=1}^{N}\sum_{p=1}^{P}\arccos(\frac{e_{n,t,p}^T\hat{e}_{n,t,p}}{\|e_{n,t,p}\|\|\hat{e}_{n,t,p}\|}) \quad (44)$$

where $a_{n,t}$ is the true abundance value of $n$th pixel at $t$ time, $\hat{a}_{n,t}$ represents the estimated abundance. $y_{n,t}$ is the original observed image of $n$th pixel at $t$ time. The reconstructed specifically at time instants $t \in \{2; 3; 4; 5\}$. Hyperspectral images were generated using the linear mixing model and white Gaussian noise was added to achieve a signal-to-noise ratio (SNR) of 30 dB.

Synthetic dataset 2 consists of a longer sequence of $T = 15$ temporal hyperspectral images. Each image has $N = 50\times50$ pixels and $L = 198$ spectral bands. This dataset contains $P = 4$ distinct endmembers (water, vegetation, image is obtained by multiplying $\hat{a}_t$ and $\hat{e}_{n;t}$, where $\hat{e}_{n;t}$ is predicted endmember.

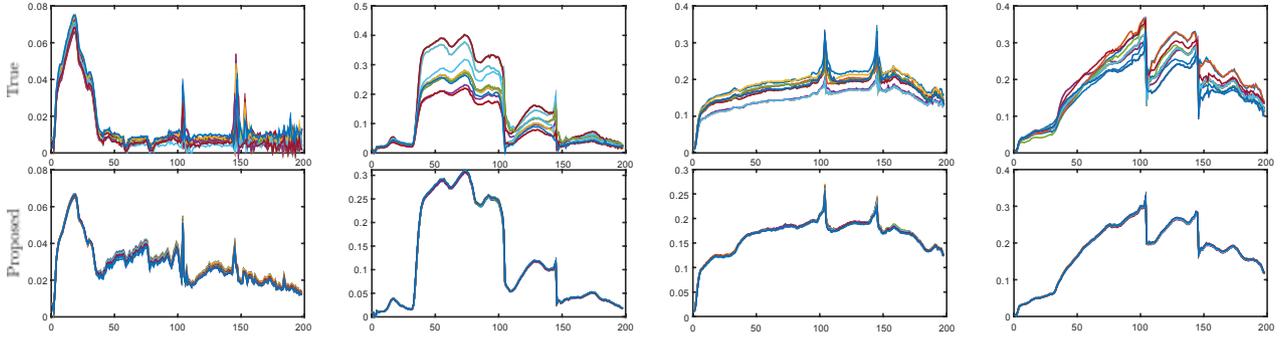

**Figure 8:** Estimated endmember on Synthetic data 2 obtained from moment 1 to moment 15. The top row represents the true endmember results and the bottom row is the result of the proposed method. Different color lines describe multiple phrases. (Left to right) Endmember 1, 2, 3 and 4.

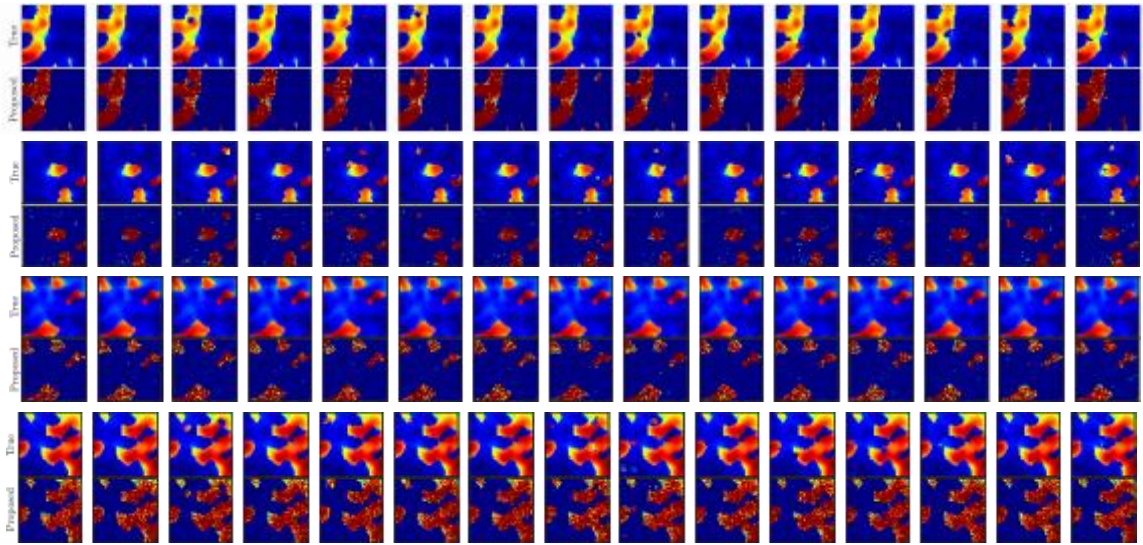

**Figure 9:** Estimated abundances of 4 different endmembers on Synthetic data 2. The top row represents the true abundance results and the bottom row is the result of the proposed method. (Top to bottom) Abundance maps for endmembers 1 to 4. (Left to right) Abundance maps obtained from moment 1 to moment 15.

### 4.2. Real Results

Here, we show the results of applying the proposed multitemporal hyperspectral unmixing approach to real-world datasets Lake Tahoe. These experiments demonstrate how the model performs when dealing with actual hyperspectral data, which may contain noise, temporal variations, and other real-world challenges. We compare the outcomes with existing methods to highlight the practical benefits of our approach.

Experimental results based on Lake Tahoe dataset reveal performance differences between methods, which can be seen in Fig.5. FCLS yielded the poorest unmixing results, exhibiting endmember confusion between soil and vegetation and missing endmembers in 5/6 phase. Kalman-based methods showed limitations in phrase 3,4 and 6 at endmember water, which are phases with small lake area. Temporal-aware methods demonstrated improved stability. OU showed similar performance to Kalman, though both still produced water misclassifications. ReSUDNN achieved clear separation, but it displayed strange material in up left corner, which also happened in MUFormer. MiLD avoid the strange phenomenon, but it is mixed at phrase 3,4 and 5 between soil and vegetation. The proposed method outperformed all benchmarks, delivering superior abundance mapping with minimal endmember confusion and effectively leveraging temporal information. This demonstrates critical importance of modeling temporal dependency for MTHU.

### 4.3. Synthetic Results

In this subsection, we present the results obtained from the synthetic dataset. The synthetic experiments are designed to test the performance of the proposed method under controlled conditions where the true abundance maps and spectral signatures are known. These results provide insight into the ability of model to accurately unmix hyperspectral data with varying degrees of complexity in terms of temporal changes.

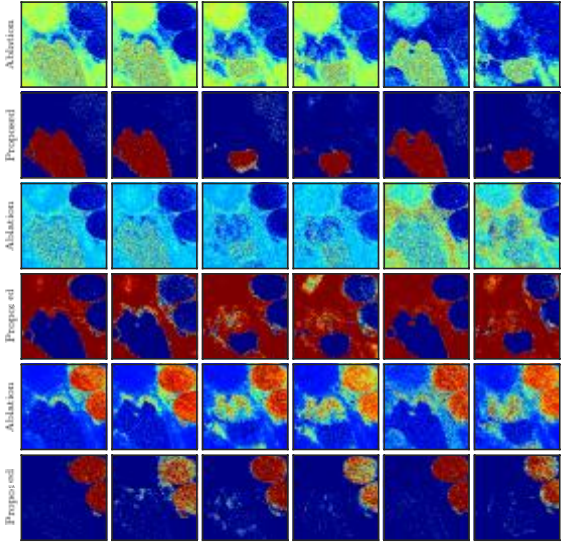

**Figure 10:** Comparison of abundance maps on Lake Tahoe. (Top to bottom) Abundance maps obtained by ablation and proposed method.(Left to right) Abundance maps obtained from moment 1 to 6.

Quantitative results (Tab.1) reveal that proposed method achieves the best abundance and endmember estimation ($NRMSE_A$; $NRMSE_E$; $SAM_E$) for synthetic dataset 1. MiLD performed best $NRMSE_Y$. Kalman filter yielded reasonable results for this dataset. FCLS performed poorly, exhibiting significant errors.

Visually results on synthetic dataset1 can be seen at Fig.6and Fig.7, which present endmembers and abundances respectively. Different from real dataset, synthetic dataset can display ground truth value. For endmember estimation (Fig.6), proposed method recovers endmembers from 1 to 3 accurately, though retrieved variability was lower than ground truth. For abundance prediction (Fig.7), OU and FCLS results appeared noisier and indicated overly mixed pixels. FCLS performed reasonably until phrase 4 but failed catastrophically at phrase 6. Kalman, ReDSUNN, MUFormer and MiLD showed suboptimal material separation. Our predicted abundances most closely resembled the ground truth, effectively capturing abrupt changes (small triangles at phrase 2-4).

For the more challenging synthetic dataset 2 (15 phases), we only display the comparison between groundtruth and proposed method. Visually, the estimated endmembers (Fig.8) and abundances (Fig.9) demonstrated excellent alignment with ground truth, effectively handling the long sequence.

### 4.4. Ablation Study

In this section, we conduct an ablation study to analyze the contributions of our copula estimation module. By systematically removing it, we evaluate how copula affects the overall performance. This experiment provides valuable insights into copula functions are crucial for improving the accuracy of hyperspectral unmixing in a multitemporal setting.

The ablation study confirms the critical role copula guidance. The full model outperforms copula-free variant on all four metrics on synthetic dataset 2, which is displayed in Tab.2. Visual results (Fig.10) demonstrate the impact of copula on dependency modeling. Non-copula configurations exhibited endmember mixture on first and third line, while copula integration produced specific improvement. The results show that copula theory fundamentally advances dependency modeling by temporal guidance.

**Table 2**
Qualiative Results of ablation study on synthetic dataset 2.

| method | $NRMSE_A$ | $NRMSE_E$ | $SAM_E$ | $NRMSE_Y$ |
|---|---|---|---|---|
| copula × | 0.5623 | 0.3107 | 0.5860 | 0.1935 |
| copula ✓ | **0.1680** | **0.1666** | **0.2004** | **0.1630** |

## 5. Conclusion

In this paper, we propose a copula guided temporal dependency method for MTHU, which formulates MTHU as new mathematical model, constructs copula-guided framework, designs two key modules with theoretical support. The mathematical model incorporates copula theory, which defines MTHU problem by utilizing copula function. The copula-guided framework estimates endmembers variability and abundances dynamics. The two modules calculates and utilizes temporal dependency into unmixing process. The theoretical support demonstrates that copula function is valid and temporal dependency exists in MTHU. The major contributions of this paper include defining MTHU problem, proposing copula-guided framework, developing key modules and providing theorical support. Our experiments on both synthetic and real datasets validates the utility of our work.

### Acknowledgement

The work was supported by the National Key Research and Development Program of China under Grant 2022YFA1003800, and the Fundamental Research Funds for the Central Universities under grant 63243074.